# Some Problems for Convex Bayesians

Henry E. Kyburg, Jr.
Department of Computer Science
University of Rochester
Rochester, NY 14627

Michael Pittarelli
Computer Science Department
SUNY Institute of Technology
Utica, NY 13504-3050

## Abstract

We discuss problems for convex Bayesian decision making and uncertainty representation. These include the inability to accommodate various natural and useful constraints and the possibility of an analog of the classical Dutch Book being made against an agent behaving in accordance with convex Bayesian prescriptions. A more general set-based Bayesianism may be as tractable and would avoid the difficulties we raise.

## 1 CONVEX BAYESIANISM

Convex Bayesianism [Levi, 1980, 1985; Snow, 1986, 1991; Stirling and Morrell, 1991] replaces the single, numerically determinate probability function required of an agent by strict Bayesians with a convex set of such functions. A convex Bayesian conditionalizes on evidence $E$ by replacing a set $S$ of probability functions with the set
$S' = \{p' \mid p'(A) = p(A \text{ and } E)/p(E), \text{ for some } p \in S\}$
If $S$ is convex, so is $S'$ [Levi, 1980]. There is no consensus among convex Bayesians regarding a decision method. The leading contender is Levi's [1980] *E-admissibility* criterion: an action is E-admissible if it maximizes expected utility relative to some probability function in the convex set.

When the set contains only one function, convex conditionalization and E-admissibility reduce to their strict Bayesian counterparts. In the usual case (qualitative probability judgments, marginals, or bounds on certain probabilities are available), the set is a polytope. For polytopes, there are linear programming methods for conditionalizing [Snow, 1991] and for determining E-admissibility [Pittarelli, 1991].

Thus, with respect to decision making and representing and updating uncertainty, convex Bayesianism includes strict Bayesianism as a special case. Further, Kyburg [1987] has demonstrated that any *belief function* over a frame of discernment $W$ corresponds to a convex set $S$ of classical probability functions over $W$, where
$$Bel(X) = \min_{p \in S} p(X),$$
but not conversely. There exist constraints determining convex sets $S$ for which a belief function cannot be found satisfying this correspondence. The set of probability functions over $W = \{w_1, ..., w_4\}$ satisfying $0.15 \leq p(w_j) \leq 0.40$ is one such set [Weichselberger and Pohlman, 1990]. (A more realistic, but more elaborate, example is discussed in [Kyburg, 1987].) Thus, all of the representational power of the belief function formalism, and more, is possessed by convex Bayesianism. As if these were not good enough reasons to convert to convex Bayesianism, even Savage regarded a convex set of probability functions as a "tempting representation of the unsure" [1972, p. 58].

So what's wrong with convex Bayesianism?

## 2 PROBLEMS

There are natural constraints on probability judgments that cannot be represented by convex sets of classical probability functions. Working with the convex hull of a nonconvex set of probability functions may result in unnecessary indecisiveness. The E-admissibility criterion is justified by the convex Bayesian's attitude that any of the probability functions in his set is permissible for use in decision making; but an analog of the classical Dutch Book can be made against an agent who replaces his nonconvex set of probability functions with a member of its convex hull. Finally, convex pooling methods have some undesirable properties.

### 2.1.1 Disjunctive Constraints

Suppose that one is informed that a die has been manufactured in such a way that either the outcome '1' is favored at the expense of '2' by 1/12 or conversely and is otherwise fair. The possible biases (either of which may be adopted as probabilities for the toss outcomes) are

$$\{(\tfrac{1}{12},\tfrac{3}{12},\tfrac{1}{6},\ldots,\tfrac{1}{6}),(\tfrac{3}{12},\tfrac{1}{12},\tfrac{1}{6},\ldots,\tfrac{1}{6})\}. \quad (*)$$

This is not a convex set. Neither is the physically more realistic

$$\{(\tfrac{1}{12}+\epsilon,\tfrac{3}{12}-\epsilon,\tfrac{1}{6},\ldots,\tfrac{1}{6}) \mid \epsilon \in [-\tfrac{1}{48},\tfrac{1}{48}]\} \cup$$
$$\{(\tfrac{3}{12}+\epsilon,\tfrac{1}{12}-\epsilon,\tfrac{1}{6},\ldots,\tfrac{1}{6}) \mid \epsilon \in [-\tfrac{1}{48},\tfrac{1}{48}]\}.$$

It may be argued that these numbers represent possible frequencies which in turn justify a *range* of permissible probabilities, with probabilities construed as *betting rates*; i.e., the frequencies (*) make acceptable any odds in the range 1:11 to 3:9 for a bet on '1'. However, while any number in this range represents reasonable odds for a single bet, we know that in the long run only one extreme or the other will represent a break-even set of odds for a sequence of bets. In the long run any odds other than one of the extremes is doomed to lead to loss.

Note that because the biases (*) are not representable by a convex set of probability functions with sample space equal to the set of six possible die toss outcomes, it follows [Kyburg, 1987] that they are not representable by a single Dempster-Shafer belief function with this set as its frame of discernment. The closest it seems possible to come is to assign probability masses

$$m(\{1\}) = m(\{2\}) = \tfrac{1}{12}, \; m(\{1,2\}) = \tfrac{1}{6},$$
$$m(\{3\}) = \cdots = m(\{6\}) = \tfrac{1}{6}.$$

But there seems to be no way to specify that the mass of $\tfrac{1}{6}$ on $\{1,2\}$ must either all go to $\{1\}$ or all go to $\{2\}$.

### 2.1.2 Independence

Judgments of irrelevance (conditional irrelevance), that is, probabilistic independence (conditional independence), are often made, are natural to make, can be made reliably, and provide well-known computational advantages [Pearl, 1988]. But the constraint of independence, combined with information that itself would determine a convex set of probability functions, will yield a nonconvex set.

The simplest imaginable example is given by Jeffrey [1987], who points out that the set of probability functions expressing the irrelevance to each other of a given pair of propositions, and nothing more, is not convex.

More generally, conditional probabilistic independence is incompatible with a convex set representation. Suppose that *interval-valued* probability distributions have been determined for the propositional variables $X$, $Y$ and $Z$ as

| $X$ | $Y$ | $i_{XY}$ | $Y$ | $Z$ | $i_{YZ}$ |
|---|---|---|---|---|---|
| $x$ | $y$ | [0,0.2] | $y$ | $z$ | [0,0.2] |
| $x$ | $\bar{y}$ | [0.1,0.3] | $y$ | $\bar{z}$ | [0.1,0.3] |
| $\bar{x}$ | $y$ | [0.2,0.4] | $\bar{y}$ | $z$ | [0.2,0.4] |
| $\bar{x}$ | $\bar{y}$ | [0.3,0.5] | $\bar{y}$ | $\bar{z}$ | [0.3,0.5] |

The set $P_{XY}$ of classical, real-valued distributions over variables $X$ and $Y$ compatible with the interval distribution $i_{XY}$ is the set of distributions $p_{XY}$ such that $p_{XY}(.)$ is a value in the interval $i_{XY}(.)$, and similarly for $P_{YZ}$. Either set is convex, the set of solutions to a system of linear inequalities. The set $P_{XYZ}$ of all classical distributions on $\{X, Y, Z\}$ whose marginals on $\{X, Y\}$ and $\{Y, Z\}$ are elements of $P_{XY}$ and $P_{YZ}$, respectively, is also a convex set. But the subset $CI_{XYZ}$ of classical distributions that in addition satisfy the constraint that variables $X$ and $Z$ are conditionally independent, given $Y$, is not convex. The distributions $p_{XYZ}$ and $p'_{XYZ}$ below are elements of $CI_{XYZ}$, but the distribution

$$q_{XYZ} = \tfrac{1}{2} p_{XYZ} + \tfrac{1}{2} p'_{XYZ}$$

is not. It fails to satisfy the conditional independence constraints:

$$q_{XYZ}(x\bar{y}z) = 0.0\bar{6} \neq 0.0625 = q_{XY}(x\bar{y}) \times q_{YZ}(\bar{y}z)/q_Y(\bar{y}).$$

| $X$ | $Y$ | $Z$ | $p_{XYZ}$ | $p'_{XYZ}$ | $q_{XYZ}$ |
|---|---|---|---|---|---|
| $x$ | $y$ | $z$ | 0.1 | 0.05 | 0.075 |
| $x$ | $y$ | $\bar{z}$ | 0.1 | 0.05 | 0.075 |
| $x$ | $\bar{y}$ | $z$ | 0.0$\bar{3}$ | 0.1 | 0.0$\bar{6}$ |
| $x$ | $\bar{y}$ | $\bar{z}$ | 0.0$\bar{6}$ | 0.1 | 0.08$\bar{3}$ |
| $\bar{x}$ | $y$ | $z$ | 0.1 | 0.15 | 0.125 |
| $\bar{x}$ | $y$ | $\bar{z}$ | 0.1 | 0.15 | 0.125 |
| $\bar{x}$ | $\bar{y}$ | $z$ | 0.1$\bar{6}$ | 0.2 | 0.18$\bar{3}$ |
| $\bar{x}$ | $\bar{y}$ | $\bar{z}$ | 0.3$\bar{3}$ | 0.2 | 0.2$\bar{6}$ |

Therefore, $CI_{XYZ}$ is not convex.

### 2.2 BETTING ON INDEPENDENT EVENTS

Suppose an agent knows that a coin is biased toward tails, but he thinks that it and its tossing mechanism yield a set of outcomes with a binomial distribution. In particular, suppose he thinks that the probability of heads on a single toss is in the interval $[0.1, 0.5]$.

It is a deductive consequence of these assumptions that the agent's beliefs regarding the four possible outcomes of a pair of tosses are bounded by the

pairs of numbers:

|      | HH   | HT   | TH   | TT   |
|------|------|------|------|------|
| low  | 0.01 | 0.09 | 0.09 | 0.25 |
| high | 0.25 | 0.25 | 0.25 | 0.81 |

Two real-valued distributions satisfying the independence and upper and lower probability constraints are:

| Toss 1 | Toss 2 | $p_1$ | $p_2$ |
|--------|--------|-------|-------|
| H | H | 0.01 | 0.25 |
| H | T | 0.09 | 0.25 |
| T | H | 0.09 | 0.25 |
| T | T | 0.81 | 0.25 |

According to convex Bayesian doctrine, any mixture (convex combination) of distributions belonging to the representation of beliefs also belongs; so the agent should include in his set the mixture
$$q = \frac{1}{2}p_1 + \frac{1}{2}p_2:$$

| Toss 1 | Toss 2 | $q$ |
|--------|--------|-----|
| H | H | 0.13 |
| H | T | 0.17 |
| T | H | 0.17 |
| T | T | 0.53 |

Observe that this set of betting odds on the outcome of a pair of tosses is perfectly coherent; this is what we expect: the mixture of a pair of coherent betting functions is coherent.

But now, as in the wiley-antagonist model of the Dutch Book Theorem [Ramsey, 1931; deFinetti, 1974], let the agent post odds in accordance with this coherent distribution of beliefs. The wiley antagonist (a) sells our agent a ticket for $13.00 that returns $100.00 if $HH$, and nothing otherwise. That's fair. Likewise, it is fair for our antagonist (b) to buy from our agent for $25.50 a ticket that returns $150.00 if $HT$ and nothing otherwise. Any collection of such bets on pairs of tosses would be considered fair.

Let us say that an agent is *booked in expectation* if, whatever the true state of the world consistent with his beliefs, his long run expectation is negative. The expectation of this set of bets is negative almost everywhere, whatever the state of the world consistent with the agent's beliefs. In the long run, our agent is almost sure to lose.

This may be seen as follows: We compute the expectation of the pair of bets (a) and (b), from the wiley fox's point of view, as a function of the probability, $p$, of heads. The payoffs for each outcome are

|     | HH | HT | TH | TT |
|-----|------|------|------|------|
| (a) | −$87.00 | $13.00 | $13.00 | $13.00 |
| (b) | −$25.50 | $124.50 | −$25.50 | −$25.50 |
| net | −$112.50 | $137.50 | −$12.50 | −$12.50 |

The expected payoff to the wiley fox is then
$$-\$250.00p^2 + \$150.00p - \$12.50.$$
This function is concave; it is positive for all values of $p \in [0.1, 0.5]$ except the endpoints, at both of which it equals $0.00, reaching a maximum of $10.00 for $p = 0.3$. The fox's expectation is positive almost everywhere, relative to what *both* the agent and the fox know. But the agent's odds are perfectly coherent.

Although the distribution $q$ is a convex combination of distributions satisfying the upper and lower probability constraints and the independence constraint, it fails to satisfy the latter. Had the agent posted odds in accordance with a distribution $q'$ exhibiting independence, the antagonist's long-run expectation (and his own), relative to $q'$, would have been zero.

To see this, let $\alpha$ and $\beta$ denote the return value of the tickets sold by the antagonist and by the agent, respectively. Let $w = q'(HH)$, $x = q'(HT)$, $y = q'(TH)$, and $z = q'(TT)$. The table of payoffs to the antagonist is

|     | HH | HT | TH | TT |
|-----|------|------|------|------|
| (a) | $-\alpha + \alpha w$ | $\alpha w$ | $\alpha w$ | $\alpha w$ |
| (b) | $-\beta x$ | $\beta - \beta x$ | $-\beta x$ | $-\beta x$ |
| net | $-\alpha + \alpha w - \beta x$ | $\beta + \alpha w - \beta x$ | $\alpha w - \beta x$ | $\alpha w - \beta x$ |

The antagonist's expectation relative to $q'$ is then
$$w(-\alpha + \alpha w - \beta x) + x(\beta + \alpha w - \beta x) +$$
$$y(\alpha w - \beta x) + z(\alpha w - \beta x),$$
which equals 0 if
$$x = \sqrt{w} \times (1-\sqrt{w}) = \sqrt{w} - w,$$
$$y = \sqrt{w} - w,$$
$$z = (1-\sqrt{w})^2,$$
i.e., if $q'$ embodies the independence constraint.

## 2.3 CONVEX POOLING AND DECISIONS WITH CONVEX HULLS

The traditional solution to the problem of pooling a set of classical probability functions is to replace them with a member of their *convex hull*, the set of all possible convex combinations of them. One may work instead with the *entire* convex hull. This is the approach of Levi [1980], who regards any of the elements of the convex hull to be a permissible resolution of the conflict among the functions and hence permissible for use in decision making.

It can affect the outcome of a decision problem if, instead of a nonconvex set $S$, its convex hull, $conv(S)$, is regarded as the set of permissible probability functions. Let $D(a)$ denote the (convex) set of probability functions relative to which action $a$ maximizes expected utility. (For distinct $a$ and $b$, $D(a)$ and $D(b)$ may share boundary points, but not interior points.) Then $a$ is E-admissible relative to a set of functions $S$ iff $D(a) \cap S \neq \emptyset$. Thus, if an action is E-admissible relative to a set $S$ of probability functions, then it is E-admissible relative to its convex hull (since $S \subseteq conv(S)$), but not conversely.

Consider the set of probability functions $S = \{p_1, p_2\}$, where
$$p_1(c_1) = \frac{1}{8}, \ p_1(c_2) = \frac{3}{4}, \ p_1(c_3) = \frac{1}{8}$$
and
$$p_2(c_1) = \frac{3}{4}, \ p_2(c_2) = \frac{1}{8}, \ p_2(c_3) = \frac{1}{8},$$
and the decision matrix

|  | $c_1$ | $c_2$ | $c_3$ |
|---|---|---|---|
| $a_1$ | 3 | 3 | 4 |
| $a_2$ | 2.5 | 3.5 | 5 |
| $a_3$ | 1 | 5 | 4 |

Relative to set $S$, only actions $a_1$ and $a_3$ are (E-)admissible; $a_2$ is not:
$$eu(a_2, p_1) = 3.5625 < 4.375 = eu(a_3, p_1)$$
$$eu(a_2, p_2) = 2.9375 < 3.125 = eu(a_1, p_2).$$
Relative to $conv(S)$, all three are admissible. (For example, $a_2$ maximizes expected utility for the probability function $\frac{1}{2}p_1 + \frac{1}{2}p_2$.)

Because any set $D(a)$ is convex, if there is consensus on an (meu-) admissible action on the part of members of a group who agree on utilities but disagree on probabilities, then that same action is admissible for any function within the convex hull of their individual probability functions. When there is not consensus, it is not obvious that a sufficient condition for admissibility of a group action is that it maximise expected utility relative to some convex combination of the group probabilities. But if any such function is permissible, then any action maximizing expected utility relative to one of these functions should be admissible. But this clashes, for example, with the intuitions behind Savage's [1972] *group minimax* criterion. The group minimax rule prescribes selection of an action "such that the largest loss faced by any member of the group will be as small as possible [p. 174]", where the loss associated with an action by a member of the group is the difference between its expectation for that member and the expectation of the action with maximum expectation for that member.

There exist decision problems for which the action maximising expectation for a convex combination of group probabilities is not a group minimax solution, although the action maximising expectation for a probability distribution outside the convex hull of the group probabilities is a group minimax solution.

Consider a group containing three members facing a decision among three actions who recognize three relevant states of nature and agree on the utility matrix above. Suppose that the group does not wish to consider mixed actions and that the group opinions are:

|  | $c_1$ | $c_2$ | $c_3$ |
|---|---|---|---|
| $p_1$ | 1/8 | 3/4 | 1/8 |
| $p_2$ | 1/4 | 1/2 | 1/4 |
| $p_3$ | 3/8 | 3/8 | 1/4 |

Then action $a_3$ is optimal (maximises expectation) for members 1 and 2, and action $a_2$ is optimal for member 3. For the mixture of distributions
$$p = \frac{1}{8}p_1 + \frac{1}{8}p_2 + \frac{3}{4}p_3,$$
action $a_2$ is optimal. But $a_3$ is the group minimax solution. It is optimal for infinitely many distributions outside $conv(\{p_1, p_2, p_3\})$, for example,
$$p'(c_1) = \frac{1}{3}, \ p'(c_2) = \frac{1}{2}, \ p'(c_3) = \frac{1}{6}.$$

Note that the group minimax criterion is stronger than Pareto-optimality. In the example above, action $a_2$, which is not a group minimax action, is Pareto-optimal since $p_3$ is an interior point of $D(a_2)$. But any group minimax action is Pareto-optimal. (Suppose it were not. Then there would exist an action for which the expectation for some agent is strictly higher and for all agents is no lower, i.e., for which the "largest loss faced by any member" is strictly less.) Pareto-optimality can also be violated by choosing a group action maximizing expectation relative to an arbitrary member of $conv(S)$ (but less easily). Let $S$ again consist of three distributions, $p_1$, $p_2$, and $p_3$. Suppose $\{p_1, p_2\} \subseteq D(a_i) \cap D(a_j)$ and $p_3 \in D(a_i) - D(a_j)$. Then $a_j$ maximizes expectation relative to the mixture $\frac{1}{2}p_1 + \frac{1}{2}p_2 + 0p_3$, but is not Pareto-optimal.

Suppose that the (classical) probability assessments of $n$ experts are to be combined to form a single probability function and that each expert assesses his probabilities in a manner consistent with the belief in the probabilistic independence of certain of the events under consideration. A convex combination of the $n$ functions will not necessarily exhibit the independence relations agreed on by the experts. If these relations are merely numerical artifacts, then the failure to preserve them is not a failure of the pooling method. On the other hand, there will be situations in which the experts each do intend for certain independence relations to hold (and may actually use them as a guide in the construction of their full probability function over the algebra of events).

Convex pooling of probability functions exhibits what may be called the *marginalization effect*. With a fixed set of pooling weights, the result of pooling a set of probability functions and then marginalizing is the same as the result of first marginalizing the functions individually and then pooling. While some regard this as a point in favor of convex pooling schemes [McConway, 1981], we do not.

*Example.* Mr. X is unsure whether the legal residence of former U. S. president Richard M. Nixon is New Jersey or California. He calls two political science professors, P and Q, neither of whom he knows personally. He asks them both to assess the probabilities of the following: Richard Nixon is an extraterrestrial and lives in New Jersey, he is an extraterrestrial and lives in California, he is not an extraterrestrial and lives in New Jersey, and he is not an extraterrestrial and lives in California. He includes the extraterrestrial question because he has the impression that many professors are eccentric; if either professor gives positive probability to Nixon's being an extraterrestrial he will not take seriously the probability given by that professor to the propositions in which he is actually interested. Suppose the assessments he receives are:

| $p_{RE}$ | | Extraterrestrial | | |
| --- | --- | --- | --- | --- |
| | | Yes | No | $p_R$ |
| Residence | NJ | 0 | 0.85 | 0.85 |
| | CA | 0 | 0.15 | 0.15 |

and

| $q_{RE}$ | | Extraterrestrial | | |
| --- | --- | --- | --- | --- |
| | | Yes | No | $q_R$ |
| Residence | NJ | 0.9 | 0 | 0.9 |
| | CA | 0.1 | 0 | 0.1 |

Given $p_{RE}$ and $q_{RE}$ and a linear pooling scheme, the choice of weights should be 1 and 0. (Whatever else Richard Nixon may be, he is certainly not an extraterrestrial.) But it would make sense for them to be something like 1/2 and 1/2 given only $p_R$ and $q_R$. However, even for weights of 1 and 0, the marginal distribution that Mr. X takes as representing the weighted pooled opinions of the two, $p_R$, remains disturbingly close to $q_R$ on any reasonable measure of the closeness of probability distributions. (On the other hand, people who believe there are extraterrestrials on Earth may be reliable trackers of celebrities they think are extraterrestrials.)

More information is available to the pooler in the original probability functions than is available in the marginals. [Lindley, 1985]. (The marginals are deducible from the original functions, but usually not conversely.) Whether as in the example the information bears on the reliability of the experts or not, there are bound to be situations in which the extra information is wasted if one adopts a pooling method exhibiting the marginalization effect.

## 3 CONCLUSION

Strict Bayesianism is too strict. It is unreasonable to demand that an agent adopt a single real-valued probability function in any and all decision-making contexts. Probabilities should be based on statistical knowledge, i.e., knowledge of frequencies; the data usually warrant only a more-or-less narrow interval of probability for each event under consideration. Even if one somehow has real-valued probabilities for certain events, they may not be those directly relevant for a particular decision problem. For example, one may have strict Bayesian marginal or conditional probabilities, from which one may infer a convex set of strict Bayesian probability functions, from which in turn bounds on the probabilities of interest may be inferred.

Convex Bayesianism, on the other hand, is not strict enough. By allowing any convex combination of strict Bayesian probability functions compatible with the available information into the set of permissible functions, incompatible probability functions may be introduced. This is most serious in the case of probabilistic independence. Independence assumptions may often reliably be based on fundamental knowledge about causality and are necessary for efficient management of uncertainty in domains of realistic size. But the convex hull of a set of functions that satisfy an independence constraint will contain functions that do not.

Replacing a not-necessarily convex set of classical probability functions with its convex hull may result in unnecessary indecisiveness; an action may be E-admissible with respect to the convex hull of a set $S$ but not relative to $S$ itself. It may also lead to financial ruin. Posting odds in accordance with a probability function not in $S$ but in the convex hull of $S$ will result in negative long-run expectation relative to the knowledge embodied in $S$.

A more general *set-based Bayesianism* is discussed elsewhere [Kyburg and Pittarelli, 1992]. A set-based Bayesian will represent uncertainty by a set of classical probability functions, as will the convex Bayesian. However, given the same data, hers will be a subset of the convex Bayesian's representation. (Hence, her representation will be more informative, by any reasonable measure of the information content of sets of probability functions.)

Determining whether or not an action is E-admissible relative to a convex or nonconvex set $S$ for which a system of constraints can be formulated (we have discussed examples only of *linear* constraints, for the convex case) requires determining whether or not a mathematical programming problem has a feasible solution; it is not necessary to identify an optimal solution. Thus, it is not clear that the computational costs associated with a more flexible set-based Bayesianism are so much greater that one should have to settle for a strictly convex alternative to strict Bayesianism.

### Acknowledgments

This work was supported in part by the National Science Foundation, the Signals Warfare Center of the U. S. Army, and the Research Foundation of the State University of New York.